%% file: acl2020_random_embeds.tex
\documentclass[11pt,a4paper]{article}
\usepackage[hyperref]{acl2020}
\newcommand{\vsp}{\vspace{0.0in}}
\newcommand{\vsppos}{\vspace{0.1in}}

\usepackage{times}
\usepackage{latexsym}

\usepackage{microtype}
\usepackage{natbib}
\usepackage{amsmath}
\usepackage{amssymb}
\usepackage{booktabs}
\usepackage{multirow}
\usepackage{makecell}
\usepackage[normalem]{ulem}
\useunder{\uline}{\ul}{}

\aclfinalcopy %
\def\aclpaperid{2006} %

\title{Contextual Embeddings: When Are They Worth It?}

\author{
	Simran Arora\thanks{\;\;Equal contribution.}, \hspace{1pt}Avner May\hspace{-0.4pt}$\overset{\,*}{\text{,}}$ Jian Zhang, Christopher R\'e\\
	Stanford Univeristy \\
	\texttt{\{simarora, avnermay, zjian, chrismre\}@cs.stanford.edu}\\
}
\date{}

\input{__macros}

\usepackage{cleveref}
\usepackage{graphicx}
\usepackage{fixltx2e}
\usepackage{enumitem}

\begin{document}
\maketitle
\begin{abstract}
We study the settings for which deep contextual embeddings (\eg, BERT) give large improvements in performance relative to classic pretrained embeddings (\eg, GloVe), and an even simpler baseline---\textit{random} word embeddings---focusing on the impact of the training set size and the linguistic properties of the task.
Surprisingly, we find that \textit{both} of these simpler baselines can \textit{match} contextual embeddings on industry-scale data, and often perform within 5 to 10\% accuracy (absolute) on benchmark tasks.
Furthermore, we identify properties of data for which contextual embeddings give particularly large gains: language containing complex structure, ambiguous word usage, and words unseen in training.
\end{abstract}

\section{Introduction}
In recent years, rich contextual embeddings such as ELMo \citep{elmo18} and BERT \citep{bert18} have enabled rapid progress on benchmarks like GLUE \citep{glue19} and have seen widespread industrial use \citep{google19}.
However, these methods require significant computational resources (memory, time) during pretraining, and during downstream task training and inference.
Thus, an important research problem is to understand when these contextual embeddings add significant value vs.\ when it is possible to use more efficient representations without significant degradation in performance.

As a first step, we empirically compare the performance of contextual embeddings with classic embeddings like word2vec \citep{word2vec13} and GloVe \citep{glove14}.
To further understand what performance gains are attributable to improved embeddings vs.\ the powerful downstream models that leverage them, we also compare with a simple baseline---\textit{fully random embeddings}---which encode no semantic or contextual information whatsoever.
Surprisingly, we find that in highly optimized production tasks at a major technology company, \textit{both} classic and random embeddings have competitive (or even slightly better!) performance than the contextual embeddings.\footnote{This aligns with recent observations from experiments with classic word embeddings at Apple \citep{re2020overton}.}$^,$\footnote{These tasks are proprietary, so we share these results anecdotally as motivation for our study.}

To better understand these results, we study the properties of NLP tasks for which contextual embeddings give large gains relative to non-contextual embeddings.
In particular, we study how the amount of training data, and the linguistic properties of the data, impact the relative performance of the embedding methods, with the intuition that contextual embeddings should give limited gains on data-rich, linguistically simple tasks.

In our study on the impact of training set size, we find in experiments across a range of tasks that the performance of the non-contextual embeddings (GloVe, random) improves rapidly as we increase the amount of training data, often attaining within 5 to 10\% accuracy of BERT embeddings when the full training set is used.
This suggests that for many tasks these embeddings could likely match BERT given sufficient data, which is precisely what we observe in our experiments with industry-scale data.
Given the computational overhead of contextual embeddings, this exposes important trade-offs between the computational resources required by the embeddings, the expense of labeling training data, and the accuracy of the downstream model.

To better understand when contextual embeddings give large boosts in performance, we identify three linguistic properties of NLP tasks which help explain when these embeddings will provide gains:

\vspace{-0.07in}
\begin{itemize}[leftmargin=*]
	\item \textbf{Complexity of sentence structure}: How interdependent are different words in a sentence? \vspace{-0.07in}
	\item \textbf{Ambiguity in word usage}: Are words likely to appear with multiple labels during training? \vspace{-0.07in}
	\item \textbf{Prevalence of unseen words}: How likely is encountering a word never seen during training? \vspace{-0.07in}
\end{itemize}
\vspace{-0.14in}

Intuitively, these properties distinguish between NLP tasks involving simple and formulaic text (\eg, assistant commands) vs.\ more unstructured and lexically diverse text (\eg, literary novels).
We show on both sentiment analysis and NER tasks that contextual embeddings perform significantly better on more complex, ambiguous, and unseen language, according to proxies for these properties.
Thus, contextual embeddings are likely to give large gains in performance on tasks with a high prevalence of this type of language.

\vsp
\section{Background}
\label{sec:background}
\vsp
We discuss the different types of word embeddings we compare in our study: contextual pretrained embeddings, %
non-contextual pretrained embeddings, %
and random embeddings; we also discuss the relative efficiency of these embedding methods, both in terms of computation time and memory (Sec.~\ref{subsec:runtime}). 

\vsppos
\noindent \paragraph{Pretrained contextual embeddings}
Recent contextual word embeddings, such as BERT~\citep{bert18} and XLNet~\citep{yang2019xlnet}, consist of multiple layers of transformers which use self-attention~\citep{vaswani2017attention}.
Given a sentence, these models encode each token into a feature vector which incorporates information from the token's context in the sentence.

\vsppos
\noindent \paragraph{Pretrained non-contextual embeddings}
Non-contextual word embeddings such as GloVe~\citep{glove14}, word2vec \citep{word2vec13}, and fastText \citep{fasttext18} encode each word in a vocabulary as a vector; intuitively, this vector is meant to encode semantic information about a word, such that similar words (\eg, synonyms) have similar embedding vectors.
These embeddings are pretrained from large language corpora, typically using word co-occurrence statistics.

\vsppos
\noindent \paragraph{Random embeddings}
In our study, we consider random embeddings (\eg, as in \citet{random16}) as a simple and efficient baseline that requires no pretraining.
Viewing word embeddings as $n$-by-$d$ matrices ($n$: vocabulary size, $d$: embedding dimension), we consider embedding matrices composed entirely of random values.
To reduce the memory overhead of storing these $n\cdot d$ random values to $O(n)$, we use circulant random matrices \citep{yu18} as a simple and efficient approach (for more details, see Appendix~\ref{app:embeddings}).\footnote{Note that one could also simply store the random seed, though this requires regenerating the embedding matrix every time it is accessed.}$^,$\footnote{We provide an efficient implementation of circulant random embedding matrices here: \url{https://github.com/HazyResearch/random_embedding}.}

\subsection{System Efficiency of Embeddings}
\label{subsec:runtime}
We discuss the computational and memory requirements of the different embedding methods, focusing on downstream task training and inference.\footnote{Pretrained contextual and non-contextual embeddings also require significant computational resources during pretraining. For example training BERT\textsubscript{BASE} takes 4 days on 16 TPU chips.}

\vsppos
\noindent \paragraph{Computation time} For deep contextual embeddings, extracting the word embeddings for tokens in a sentence requires running inference through the full network, which takes on the order of 10 ms on a GPU. 
Non-contextual embeddings (\eg, GloVe, random) require negligible time ($O(d)$) to extract an embedding vector.

\vsppos
\noindent \paragraph{Memory}
Using contextual embeddings for downstream training and inference requires storing all the model parameters, as well as the model activations during training if the embeddings are being fine-tuned (\eg, 440 MB to store BERT\textsubscript{BASE} parameters, and on the order of 5-10 GB to store activations).
Pretrained non-contextual embeddings (\eg, GloVe) require $O(nd)$ to store a $n$-by-$d$ embedding matrix (\eg, 480 MB to store a 400k by 300 GloVe embedding matrix).
Random embeddings take $O(1)$ memory if only the random seed is stored, or $O(n)$ if circulant random matrices are used (\eg, 1.6 MB if $n=400$k).

\vsp
\section{Experiments}
\label{sec:experiments}
\vsp
We provide an overview of our experimental protocols (Section~\ref{sec:details}), the results from our study on the impact of training set size (Section~\ref{sec:data}), and the results from our linguistic study (Section~\ref{sec:linguistics}).
We show that the gap between contextual and non-contextual embeddings often shrinks as the amount of data increases, and is smaller on language that is \textit{simpler} based on linguistic criteria we identify.

\vsp
\subsection{Experimental Details}
\label{sec:details}
To study the settings in which contextual embeddings give large improvements, we compare them to GloVe and random embeddings across a range of named entity recognition (NER) \citep{tjongkimsang2003conll}, sentiment analysis \citep{kim14}, and natural language understanding \citep{glue19} tasks. %
We choose these lexically diverse tasks as examples of word, sentence, and sentence-pair classification tasks, respectively.
For our embeddings, we consider 768-dimensional pretrained BERT\textsubscript{BASE} word embeddings, 300-dimensional publicly available GloVe embeddings, and 800-dimensional random circulant embeddings.
We keep the embedding parameters fixed during training for all embedding types (no fine-tuning), to isolate the benefits of pretraining from the benefits of task training.
We use a CNN model \citep{kim14} for sentiment analysis and a BiLSTM \citep{Akbik2018ContextualSE,glue19} for the NER and General Language Understanding Evaluation (GLUE) tasks.
For more details on the tasks, models, and training protocols, please see Appendix~\ref{app:exp_details}.

\begin{figure}
	\begin{tabular}{c c}
		\hspace{-0.2in} \includegraphics[width=0.52\linewidth]{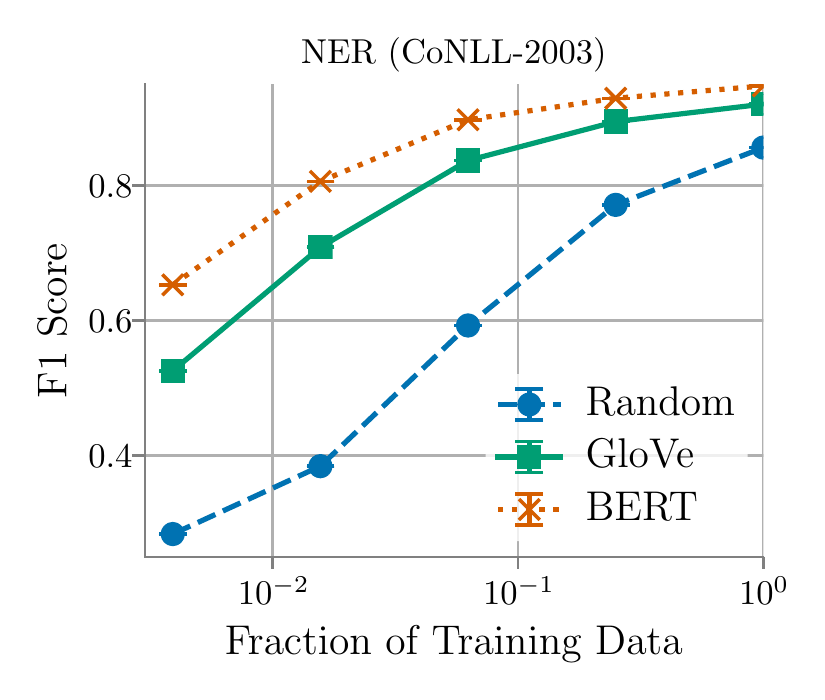} &
		\hspace{-0.3in} \includegraphics[width=0.52\linewidth]{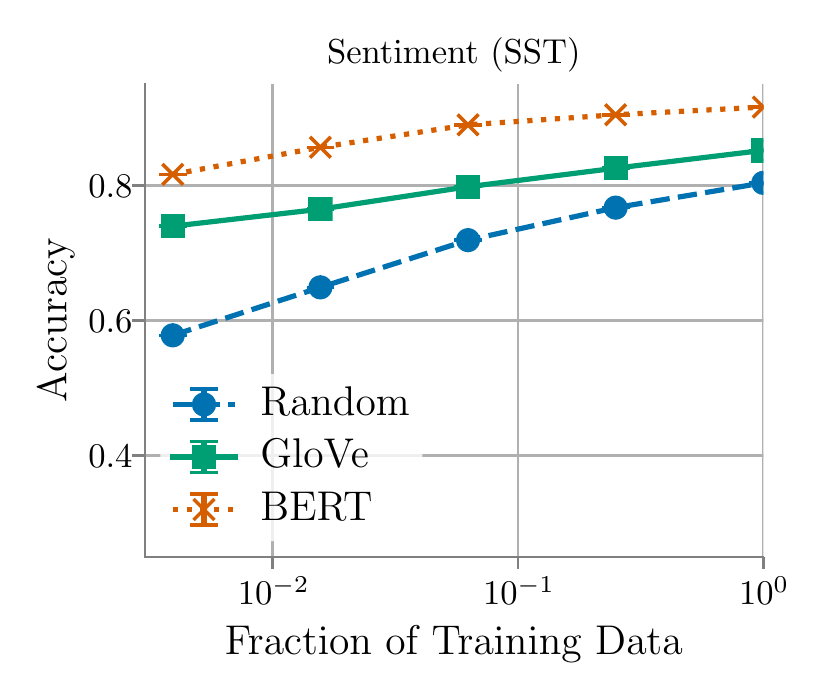}
	\end{tabular}
	\caption{NER (CoNLL-2003; left), and sentiment analysis (SST; right) performance, as a function of the fraction of the training set used.
		As the amount of training data increases, the non-contextual embedding performance improves quickly, generally narrowing the gap with the contextual embeddings.
		\hspace{\textwidth}}
	\label{fig:vary_data}
	\vsp \vsp \vsp \vsp
\end{figure}

\begin{table}[t]
	\centering
	\small
	\begin{tabular}{@{\hskip1pt}l@{\hskip3pt}l@{\hskip1pt}|l l l l@{\hskip0pt}c c@{\hskip-2pt}}
		\hline
		 \multicolumn{2}{c|}{\multirow{3}{*}{\textbf{Task}}} & \multicolumn{3}{c}{\multirow{2}{*}{\textbf{Performance gap}}} & \multicolumn{2}{|c}{\textbf{\!\!Sample}} 
		 \\
		 & & & & & \multicolumn{2}{|c}{\textbf{\!\!complexity ratio}} \\
			\cline{3-5} \cline{6-7}
		 & & \textbf{B} & \textbf{R-B} & \multicolumn{1}{l}{\textbf{G-B}} & \multicolumn{1}{|l}{\textbf{R/B}} & \textbf{G/B}\\ \hline \hline
		\textbf{NER} & CoNLL & 94.8 & -9.1 & -2.7 & \multicolumn{1}{|l}{16} & 4 \\
		\hline \hline
		\multirow{6}{*}{\textbf{Sent.}} & TREC & 95.8 & -10.3 & -6.0 & \multicolumn{1}{|l}{4} & 4 \\
		& MPQA & 89.6 & -4.7 & -0.9 & \multicolumn{1}{|l}{16} & 1 \\
		& CR & 88.5 & -5.0 & -3.5 & \multicolumn{1}{|l}{16} & 4 \\
		& SUBJ & 97.7 & -8.7 & -3.3 & \multicolumn{1}{|l}{256} & 16 \\
		& SST & 91.6 & -11.2 & -6.4 & \multicolumn{1}{|l}{256} & 64 \\
		& MR & 85.9 & -13.4 & -6.9 & \multicolumn{1}{|l}{256} & 16 \\
		\hline \hline
		\multirow{7}{*}{\textbf{GLUE}} & RTE & 61.0 & -1.8 & -2.5 & \multicolumn{1}{|l}{1} & 64 \\
		& MRPC & 84.8 & -8.8 & -6.9 & \multicolumn{1}{|l}{16} & 4 \\
		& QQP & 86.5 & -9.2 & -6.8 & \multicolumn{1}{|l}{16} & 16 \\
		& CoLA & 51.7 & -34.6 & -40.1 & \multicolumn{1}{|l}{64} & 256 \\
		& STS-B & 85.6 & -29.1 & -19.8 & \multicolumn{1}{|l}{64} & 16 \\
		& QNLI & 84.6 & -15.8 & -9.5 & \multicolumn{1}{|l}{64} & 16 \\
		& MNLI & 78.8 & -17.1 & -12.0 & \multicolumn{1}{|l}{64} & 16 \\
		& SST & 91.3 & -15.9 & -12.8 & \multicolumn{1}{|l}{256} & 64 \\
		\hline
	\end{tabular}
	\caption{\label{tab:scores} Performance and sample complexity of random (R) and GloVe (G) relative to BERT (B) for NER, sentiment analysis (Sent.), and language understanding (GLUE) tasks. %
	Second column shows BERT accuracy; third/fourth columns show the accuracy gap between BERT and random/GloVe; fifth/sixth columns show \textit{sample complexity ratios}, the largest $n\in \{1,4,16,64,256\}$ for which BERT outperforms random/GloVe when trained on $n$-times less data.
	We observe that non-contextual embeddings can often (1) perform within $10\%$ absolute accuracy of the contextual embeddings, and (2) match the performance of contextual embeddings which are trained on 1x-16x less data. 
	This sheds light on a tradeoff between the upfront cost of labeling training data and the inference-time computational cost of the embeddings.
	}
\end{table}

\vsp
\subsection{Impact of Training Data Volume}
\label{sec:data}
We show that the amount of downstream training data is a critical factor in determining the relative performance of contextual vs.\ non-contextual embeddings.
In particular, we show in representative tasks in Figure~\ref{fig:vary_data} that the performance of the non-contextual embedding models improves quickly as the amount of training data is increased (plots for all tasks in Appendix~\ref{app:training_data}).\footnote{We provide theoretical support for why random embeddings perform strongly given sufficient data in Appendix~\ref{app:theory}.}
As a result of this improvement, we show in Table~\ref{tab:scores} that across tasks when the full training set is used, the non-contextual embeddings can often (1) perform within 10\% absolute accuracy of the contextual embeddings, and (2) match the performance of the contextual embeddings trained on 1x-16x less data, while also being orders of magnitude more computationally efficient.
In light of this, ML practitioners may find that for certain real-world tasks the large gains in efficiency are well worth the cost of labeling more data.

Specifically, in this table we show for each task the difference between the accuracies attained by BERT vs. GloVe and random (note that random sometimes beats GloVe!), as well as the largest integer $n \in \{1,4,16,64,256\}$ such that BERT trained on $\frac{1}{n}$ of the training set still outperforms non-contextual embeddings trained on the full set.

\vsp
\subsection{Study of Linguistic Properties}
\label{sec:linguistics}
\vsp

In this section, we aim to identify properties of the language in a dataset for which contextual embeddings perform particularly well relative to non-contextual approaches.
Identifying such properties would allow us to determine whether a new task is likely to benefit from contextual embeddings.

As a first step in our analysis, we evaluate the different embedding types on the  GLUE Diagnostic Dataset~\citep{glue19}. 
This task defines four categories of linguistic properties;
we observe that the contextual embeddings performed similarly to the non-contextual embeddings for three categories, and significantly better for the \textit{predicate-argument structure} category (Matthews correlation coefficients of .33, .20, and .20 for BERT, GloVe, and random, respectively. See Appendix~\ref{app:glue_diagnostic} for more detailed results).
This category requires understanding how sentence subphrases are composed together (\eg, prepositional phrase attachment, and identifying a verb's subject and object). 
Motivated by the observation that contextual embeddings are systematically better on specific types of linguistic phenomena, we work to identify simple and quantifiable properties of a downstream task's language which correlate with large boosts in performance from contextual embeddings.

In the context of both word-level (NER) and sentence-level (sentiment analysis) classification tasks, we define metrics that measure (1) the complexity of text structure, (2) the ambiguity in word usage, and (3) the prevalence of unseen words (Section~\ref{sec:metrics}), and then show that contextual embeddings attain significantly higher accuracy than non-contextual embeddings on inputs with high metric values (Section~\ref{sec:metric_validation}, Table~\ref{tab:metrics}).

\subsubsection{Metric Definitions}
\label{sec:metrics}
We now present our metric definitions for NER and sentiment analysis, organized by the above three properties (detailed definitions in Appendix~\ref{app:linguistic}).

\vsppos
\noindent \paragraph{Complexity of text structure}
\label{sec:complex}
We hypothesize that language with more complex internal structure will be harder for non-contextual embeddings. We define the metrics as follows:

\begin{itemize}[topsep=0.4em, partopsep=0.0em, itemsep=0em, parsep=0.3em, leftmargin=*]
	\item \textbf{NER}: We consider the number of tokens spanned by an entity as its complexity metric (\eg, ``George Washington'' spans 2 tokens), as correctly labeling a longer entity requires understanding the \textit{relationships} between the different tokens in the entity name.
	\item \textbf{Sentiment analysis}: We consider the average distance between pairs of dependent tokens in a sentence's dependency parse as a measure of the sentence's complexity, as long-range dependencies are typically a challenge for NLP systems.
\end{itemize}

\vsppos
\noindent \paragraph{Ambiguity in word usage}
\label{sec:ambiguous}
We hypothesize that non-contextual embeddings will perform poorly in disambiguating words that are used in multiple different ways in the training set.
We define the metrics as follows:

\begin{itemize}[topsep=0.4em, partopsep=0.0em, itemsep=0em, parsep=0.3em, leftmargin=*]
	\item \textbf{NER}: We consider the number of labels (person, location, organization, miscellaneous, other) a token appears with in the training set as a measure of its ambiguity (\eg, ``Washington'' appears as a person, location, and organization in CoNLL-2003).
	\item \textbf{Sentiment analysis}: As a measure of a sentence's ambiguity, we take the average over the words in the sentence of the probability that the word is positive in the training set, and compute the entropy of a coin flip with this probability.\footnote{For
		sentiment tasks with $C$-labels ($C=6$ for the TREC dataset), we consider the entropy of the average label distribution $\frac{1}{n}\sum_{i=1}^n p(y|w_i) \in \RR^C$ over the sentence words $w_i$.
	}
\end{itemize}

\vsppos
\noindent \paragraph{Prevalence of unseen words}
\label{sec:unseen}
We hypothesize that contextual embeddings will perform significantly better than non-contextual embeddings on words which do not appear at all in the training set for the task.
We define the following metrics:

\begin{itemize}[topsep=0.4em, partopsep=0.0em, itemsep=0em, parsep=0.3em, leftmargin=*]
	\item \textbf{NER}: For a token in the NER input, we consider the inverse of the number of times it was seen in the training set (letting $1/0 \defeq \infty$).
	\item \textbf{Sentiment analysis}: Given a sentence, we consider as our metric the fraction of words in the sentence that were never seen during training.
\end{itemize}

\subsubsection{Empirical validation of metrics}
\label{sec:metric_validation}
In Table~\ref{tab:metrics} we show that for each of the metrics defined above, the accuracy gap between BERT and random embeddings is larger on inputs for which the metrics are large.
In particular, we split each of the task validation sets into two halves, with points with metric values below the median in one half, and above the median in the other.
We see that in 19 out of 21 cases, the accuracy gap between BERT and random embeddings is larger on the slice of the validation set corresponding to large metric values, validating our hypothesis that contextual embeddings provide important boosts in accuracy on these points.

In Appendix \ref{app:linguistic_glove_vs_bert}, we present a similar table comparing the performance of BERT and GloVe embeddings.
We see that the gap between GloVe and BERT errors is larger above the median than below it in 11 out of 14 of the complexity and ambiguity results, which is consistent with our hypothesis that context is helpful for structurally complex and ambiguous language.
However, we observe that GloVe and BERT embeddings---which can both leverage pretrained knowledge about unseen words---perform relatively similarly to one another above and below the median for the unseen metrics.

\begin{table}
	\centering
	\small
	\begin{tabular}{l l l l l l l}
		\hline \multirow{2}{*}{} & \multicolumn{2}{c}{\textbf{Complexity}} & \multicolumn{2}{c}{\textbf{Ambiguity}} & \multicolumn{2}{c}{\textbf{Unseen}} \\
		\cmidrule(lr){2-3}
		\cmidrule(lr){4-5}
		\cmidrule(lr){6-7}
		\!\!\!\!\textbf{Task} & Abs.\!\!\!\! & Rel.\!\!\!\! & Abs.\!\!\!\! & Rel.\!\!\!\! & Abs.\!\!\!\! & Rel.\!\!\!\! \\ \hline
		\!\!\!\!NER (CoNLL) & +4.6 & 1.4 & +7.7 & 2.0 & +5.0 & 1.4 \\
		\!\!\!\!Sent. (MR) & -5.4 & 0.7 & +3.3 & 1.3 & +1.2 & 1.1 \\
		\!\!\!\!Sent. (SUBJ) & -1.8 & 0.8 & +6.7 & 2.3 & +0.9 & 1.1 \\
		\!\!\!\!Sent. (CR) & +0.6 & 1.1 & +3.0 & 1.8 & +4.1 & 2.4 \\
		\!\!\!\!Sent. (SST) & +7.4 & 2.1 & +8.7 & 2.4 & +2.3 & 1.2 \\
		\!\!\!\!Sent. (TREC) & +5.1 & 1.7 & +5.9 & 1.8 & +4.4 & 1.5 \\
		\!\!\!\!Sent. (MPQA) & +7.9 & 13.5 & +7.1 & 12.4 & +1.3 & 1.4 \\
		\hline
	\end{tabular}
	\caption{\label{tab:metrics} For our complexity, ambiguity, and unseen prevalence metrics, we slice the validation set using the median metric value, and compute the average error rates for BERT and random on each slice.
		We show that the gap between BERT and random errors is larger on the slice above the median than below it in 19 out of 21 cases, in absolute (Abs.) and relative (Rel.) terms.
	}
	\vsp \vsp 
\end{table}

\section{Related Work}
\label{sec:relwork}

The original work on ELMo embeddings \citep{elmo18} showed that the gap between contextual and non-contextual embeddings narrowed as the amount of training data increased.
Our work builds on these results by additionally comparing with random embeddings, and by studying the linguistic properties of tasks for which the contextual embeddings give large gains.

Our work is not the first to study the downstream performance of embeddings which do not require any pretraining.
For example, in the context of neural machine translation (NMT) it is well-known that randomly-initialized embeddings can attain strong performance \citep{wu16,vaswani2017attention};
the work of \citet{qi18} empirically compares the performance of pretrained and randomly-initialized embeddings across numerous languages and dataset sizes on NMT tasks, showing for example that the pretrained embeddings typically perform better on similar language pairs, and when the amount of training data is small (but not too small).
Furthermore, as mentioned in Section~\ref{sec:background}, random embeddings were considered as a baseline by \citet{random16}, to better understand the gains from using generic vs.\ domain-specific word embeddings for text classification tasks.
In contrast, our goal for using random embeddings in our study was to help clarify when and why pre-training gives gains, and to expose an additional operating point in the trade-off space between computational cost, data-labeling cost, and downstream model accuracy.

\section{Conclusion}
\label{sec:conclusion}
We compared the performance of contextual embeddings with non-contextual pretrained embeddings and with an even simpler baseline---random embeddings.
We showed that these non-contextual embeddings perform surprisingly well relative to the contextual embeddings on tasks with plentiful labeled data and simple language.
While much recent and impressive effort in academia and industry has focused on improving state-of-the-art performance through more sophisticated, and thus increasingly expensive, embedding methods, this work offers an alternative perspective focused on realizing the trade-offs involved when choosing or designing embedding methods. 
We hope this work inspires future research on better understanding the differences between embedding methods, and on designing simpler and more efficient models.

\section*{Acknowledgments}
We gratefully acknowledge the support of DARPA under Nos. FA87501720095 (D3M), FA86501827865 (SDH), and FA86501827882 (ASED); NIH under No. U54EB020405 (Mobilize), NSF under Nos. CCF1763315 (Beyond Sparsity), CCF1563078 (Volume to Velocity), and 1937301 (RTML);  ONR under No. N000141712266 (Unifying Weak Supervision); the Moore Foundation, NXP, Xilinx, LETI-CEA, Intel, IBM, Microsoft, NEC, Toshiba, TSMC, ARM, Hitachi, BASF, Accenture, Ericsson, Qualcomm, Analog Devices, the Okawa Foundation, American Family Insurance, Google Cloud, Swiss Re, the Stanford Graduate Fellowship in Science and Engineering, and members of the Stanford DAWN project: Teradata, Facebook, Google, Ant Financial, NEC, VMWare, and Infosys. The U.S. Government is authorized to reproduce and distribute reprints for Governmental purposes notwithstanding any copyright notation thereon. Any opinions, findings, and conclusions or recommendations expressed in this material are those of the authors and do not necessarily reflect the views, policies, or endorsements, either expressed or implied, of DARPA, NIH, ONR, or the U.S. Government.

\bibliography{ref}
\bibliographystyle{acl_natbib}

\appendix
\include{appendix}

\end{document}

%% file: __macros.tex
\usepackage{amsmath}

\DeclareMathOperator{\circulant}{circ}

\newcommand{\by}{\bar{y}}

\newcommand{\ie}{i.e.}
\newcommand{\eg}{e.g.}

\def\ddefloop#1{\ifx\ddefloop#1\else\ddef{#1}\expandafter\ddefloop\fi}
\def\ddef#1{\expandafter\def\csname bb#1\endcsname{\ensuremath{\mathbb{#1}}}}
\ddefloop ABCDEFGHIJKLMNOPQRSTUVWXYZ\ddefloop
\def\ddef#1{\expandafter\def\csname c#1\endcsname{\ensuremath{\mathcal{#1}}}}
\ddefloop ABCDEFGHIJKLMNOPQRSTUVWXYZ\ddefloop

\newcommand{\RR}{\ensuremath{\bbR}} %

\newcommand{\ProbOpr}[1]{\mathbb{#1}}
\newcommand{\expect}[2]{%
\ifthenelse{\equal{#2}{}}{\ProbOpr{E}_{#1}}
{\ifthenelse{\equal{#1}{}}{\ProbOpr{E}\left[#2\right]}{\ProbOpr{E}_{#1}\left[#2\right]}}} %
\newcommand{\var}[2]{%
\ifthenelse{\equal{#2}{}}{\ProbOpr{VAR}_{#1}}
{\ifthenelse{\equal{#1}{}}{\ProbOpr{VAR}\left[#2\right]}{\ProbOpr{VAR}_{#1}\left[#2\right]}}} 

\newcommand{\diag}{\operatornamewithlimits{diag}}

\newcommand{\defeq}{:=}

\renewcommand{\paragraph}[1]{\textbf{#1}\quad}

%% file: appendix.tex
\section{Experimental Details}
\label{app:exp_details}

We now describe the embeddings (Appendix~\ref{app:embeddings}), tasks (Appendix~\ref{app:tasks}), and models (Appendix~\ref{app:downstream_models}) we use in our experiments in more detail.

\subsection{Embeddings}
\label{app:embeddings}
We compare the performance of BERT contextual embeddings with GloVe  embeddings and random embeddings.
We specifically use 768-dimensional BERT\textsubscript{BASE} WordPiece embeddings, 300-dimensional GloVe embeddings, and 800-dimensional random embeddings.
We freeze each set of embeddings prior to training, and do not fine-tune the embeddings during training.
The random embeddings are normalized to have the same Frobenius norm as the GloVe embeddings.
We now describe how we use circulant matrices to reduce the memory requirement for the random embeddings.

\paragraph{Circulant Random Embeddings}
To store a random $n$-by-$d$ matrix in $O(n)$ memory instead of $O(nd)$, we use random circulant matrices \citep{yu18}.
Specifically, we split the $n$-by-$d$ matrix into $\frac{n}{d}$ disjoint $d$-by-$d$ sub-matrices (assuming for simplicity that $d$ divides $n$ evenly), where each sub-matrix is equal to $CD$, where $C =\circulant(c) \in \RR^{d\times d}$ is a circulant matrix based on a random Gaussian vector $c \in \RR^d$, and $D = \diag(r) \in \RR^{d \times d}$ is a diagonal matrix based on a random Radamacher vector $r\in\{-1,+1\}^d$.
Note that a circulant matrix $\circulant(c)$ is defined as follows:
\begin{equation*}
\circulant(c) \defeq
\begin{pmatrix}
\scriptsize
c_0 & c_d & \dots & c_{2} & c_{1} \\
c_1 & c_0 & \dots & c_{3} & c_{2} \\
\dots & \dots & \ddots & \dots & \dots \\
c_{d-1} & c_{d-2} & \dots & c_{0} & c_{d} \\ 
c_d & c_{d-1} & \dots & c_{1} & c_{0} 
\end{pmatrix}.
\end{equation*}
Random circulant embeddings have been used in the kernel literature to make kernel approximation methods more efficient~\citep{yu2015compact}.
For downstream training and inference, one can simply store the $d$-dimensional $c$ and $r$ vectors for each of the $\frac{n}{d}$ disjoint $d$-by-$d$ sub-matrices, taking a total of $O(n)$ memory.
Alternatively, one can simply store a single random seed ($O(1)$ memory), and these $c,r$ vectors can be regenerated on the fly each time a row of the embedding matrix is accessed.
Note that in addition to being very memory efficient, random embeddings avoid the expensive pretraining process over a large language corpus.

\subsection{Tasks}
\label{app:tasks}
We perform evaluations on three types of standard downstream NLP tasks: named entity recognition (NER), sentiment analysis, and natural language understanding.
NER involves classifying each token in the input text as an entity or a non-entity, and further classifying the entity type for identified entities.
We evaluate on the CoNLL-2003 benchmark dataset, in which each token is assigned a label of ``O'' (non-entity), ``PER'' (person), ``ORG'' (organization), ``LOC'' (location), or ``MISC'' (miscellaneous).
Sentiment analysis involves assigning a classification label at the sentence level corresponding to the sentiment of the sentence.
We evaluate on five binary sentiment analysis benchmark datasets including MR, MPQA, CR, SST, and SUBJ.
We also evaluate on the benchmark TREC dataset, which assigns one of six labels to each input example.
For natural language understanding, we use the standard GLUE benchmark tasks, and the GLUE diagnostic task. 

\subsection{Downstream Task Models}
\label{app:downstream_models}
We use the following models and training protocols for the NER, sentiment analysis, and GLUE tasks:

\textbf{NER}: We use a BiLSTM task model with a CRF decoding layer, and we use the default hyperparameters from the flair \citep{flair19} repository:\footnote{\url{https://github.com/zalandoresearch/flair}.}
$256$ hidden units, $32$ batch size, $150$ max epochs, and a stop-condition when the learning rate decreases below $0.0001$ with a decay constant of $0.5$ and patience of $4$.
In our evaluation, we report micro-average F1-scores for this task.

\textbf{Sentiment analysis}: We use the architecture and training protocol from \citet{kim14}, using a CNN with $1$ convolutional layer, $3$ kernel sizes in $\{3, 4, 5\}$, $100$ kernels, $32$ batch size, $100$ max epochs, and a constant learning rate.
We report the validation error rates in evaluations of each task.

\textbf{GLUE}: We use the Jiant \citep{jiant19} implementation of a BiLSTM with $1024$ hidden dimensions, $2$ layers, $32$ batch size, and a stop-condition when the learning rate decreases below $0.000001$ with a decay constant of $0.5$ and patience of $5$.
We consider the following task-specific performance metrics: Matthews correlation for CoLA, MNLI, and the diagnostic task, validation F1-score for MRPC and QQP, and validation accuracy for QNLI and RTE. 

\section{Impact of Training Data Volume}
\label{app:training_data}
We now provide additional details regarding our experiments on the impact of training set size on performance (Appendix~\ref{app:experiment_details_data}), our complete set of empirical results from these experiments (Appendix~\ref{app:extended_results_data}), as well as theoretical support for the strong performance  of random embedding models in these experiments, when trained with sufficient downstream data (Appendix~\ref{app:theory}).

\subsection{Additional Experiment Details}
\label{app:experiment_details_data}
For each task, we evaluate performance using five fractions of the full training dataset, to understand how the amount of training data affects performance: $\{\frac{1}{4^4},\frac{1}{4^3},\frac{1}{4^2},\frac{1}{4^1},1\}$. 
For each fraction $c$, we randomly select a subset of the training set of the corresponding size, and replicate this data $1/c$ times;
we then train models using this redundant dataset, using the model architectures and training protocols described in Appendix~\ref{app:downstream_models}.
In downstream training we perform a seperate hyperparameter sweep of the learning rate at each fraction of the training data, and select the best learning rate for each embedding type.
We use the following lists of learning rates for the different tasks:
\begin{itemize}
	\item \textbf{NER}: $\{.003, .01, .03, .1, .3, 1, 3\}$.
	\item \textbf{Sentiment analysis}: $\{$1e-5, 3e-5, 1e-4, 3e-4, 1e-3, 3e-2, 1e-2$\}$.
	\item \textbf{GLUE}: $\{$1e-6, 3e-6, 1e-5, 3e-5, 1e-4, 3e-4, 1e-3$\}$. 
\end{itemize}

\subsection{Extended Results}
\label{app:extended_results_data}
\begin{figure*}
	\begin{tabular}{c}
		\hspace{1.4in}\includegraphics[width=0.45\linewidth]{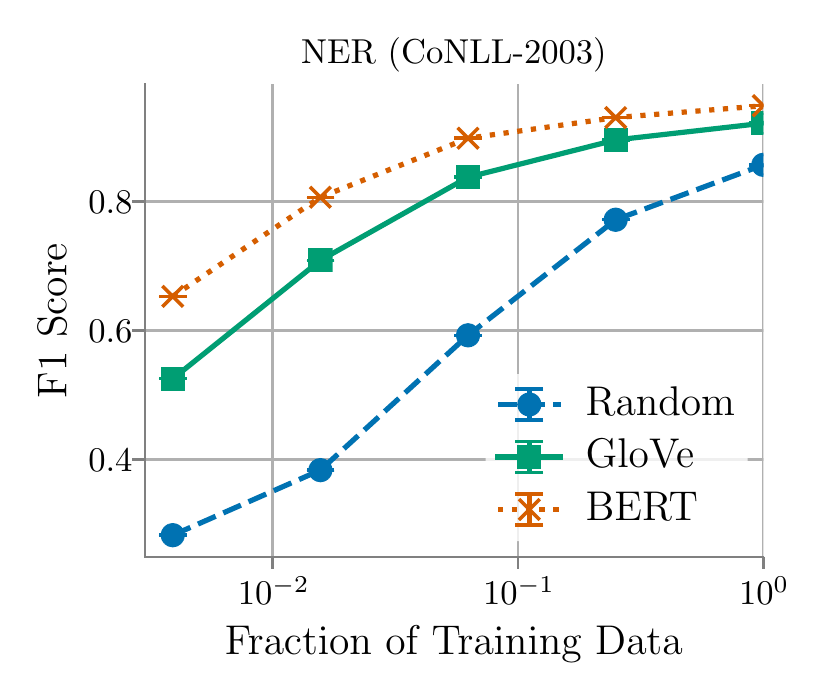}
	\end{tabular}
	\begin{tabular}{c c}
		\includegraphics[width=0.45\linewidth]{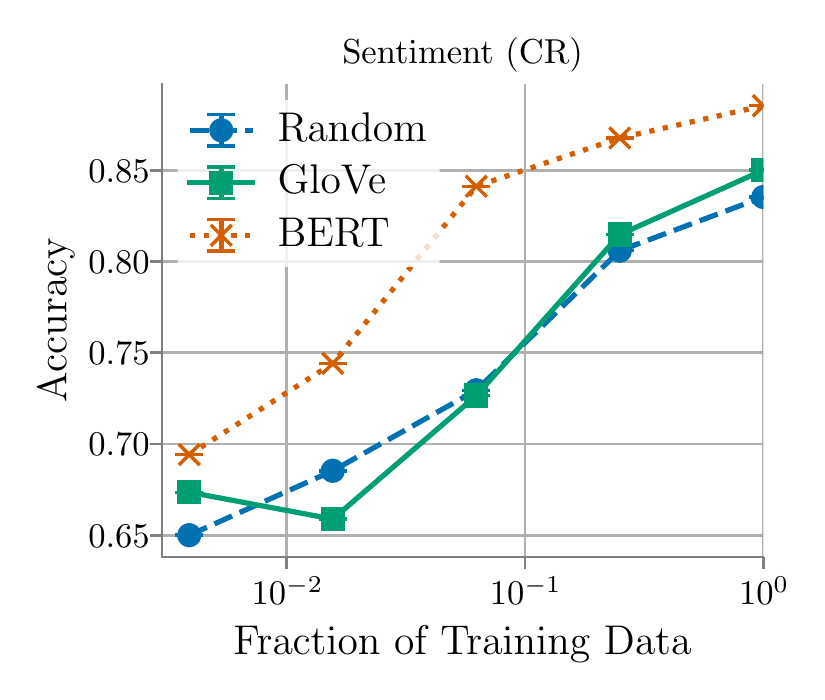} &
		\includegraphics[width=0.45\linewidth]{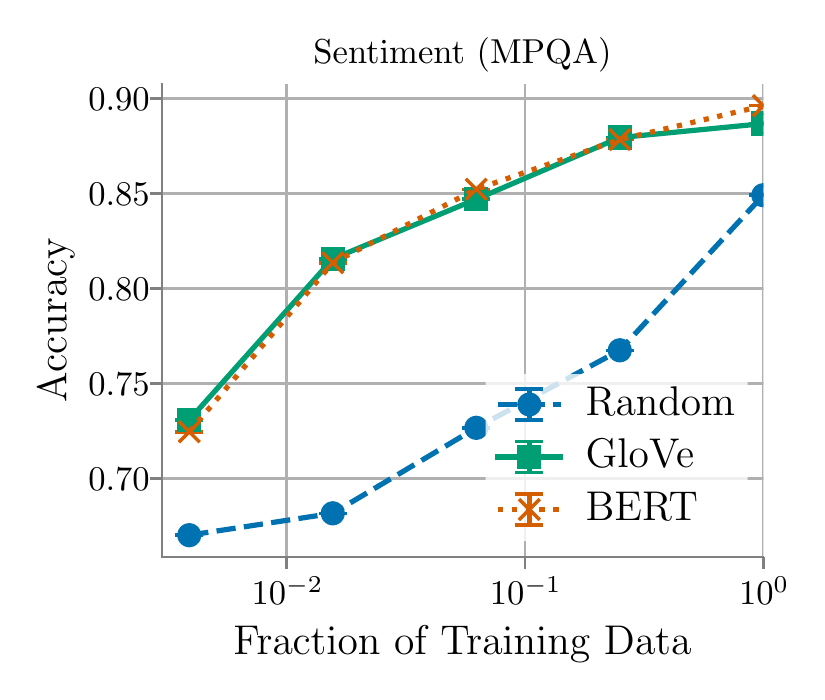} \\
		\includegraphics[width=0.45\linewidth]{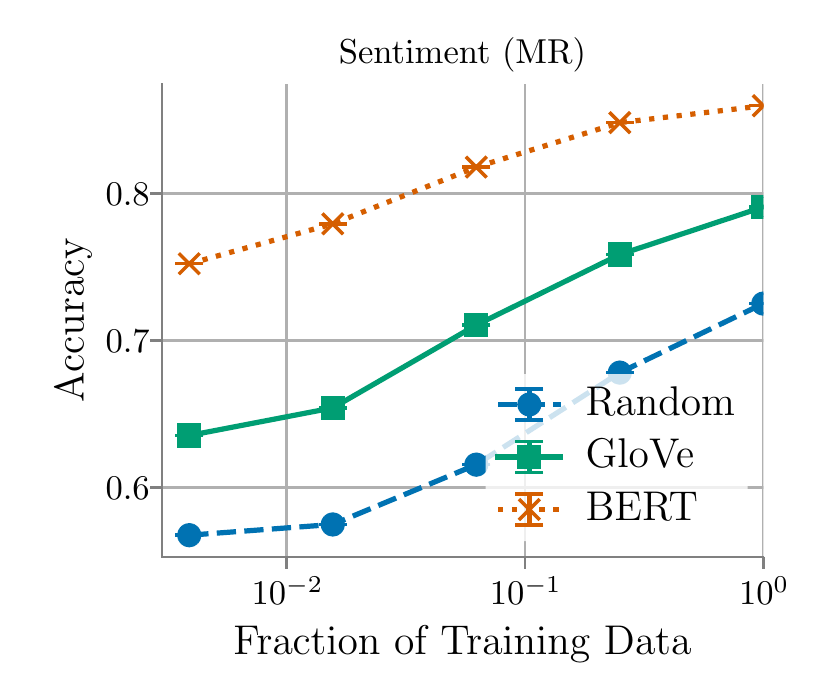} &
		\includegraphics[width=0.45\linewidth]{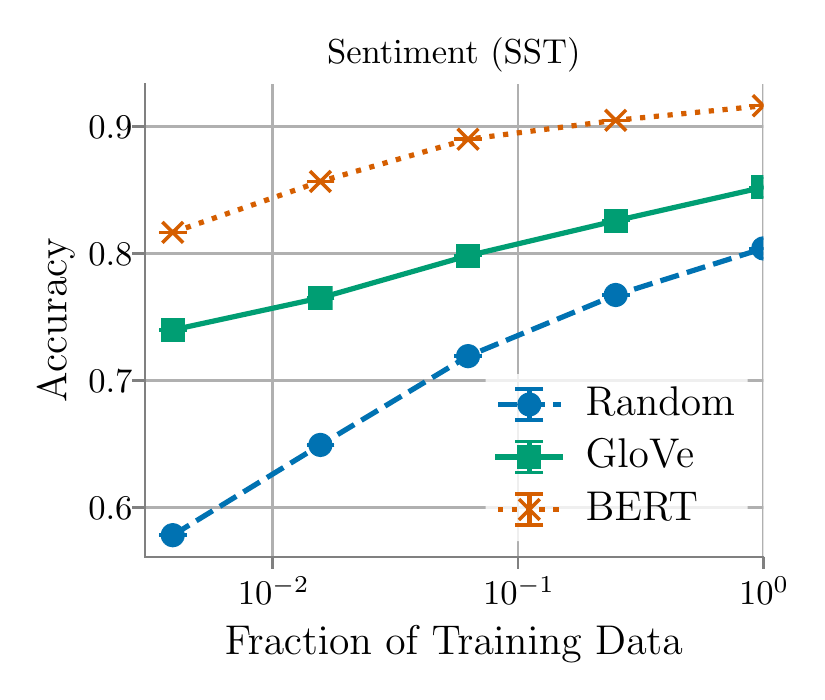} \\
		\includegraphics[width=0.45\linewidth]{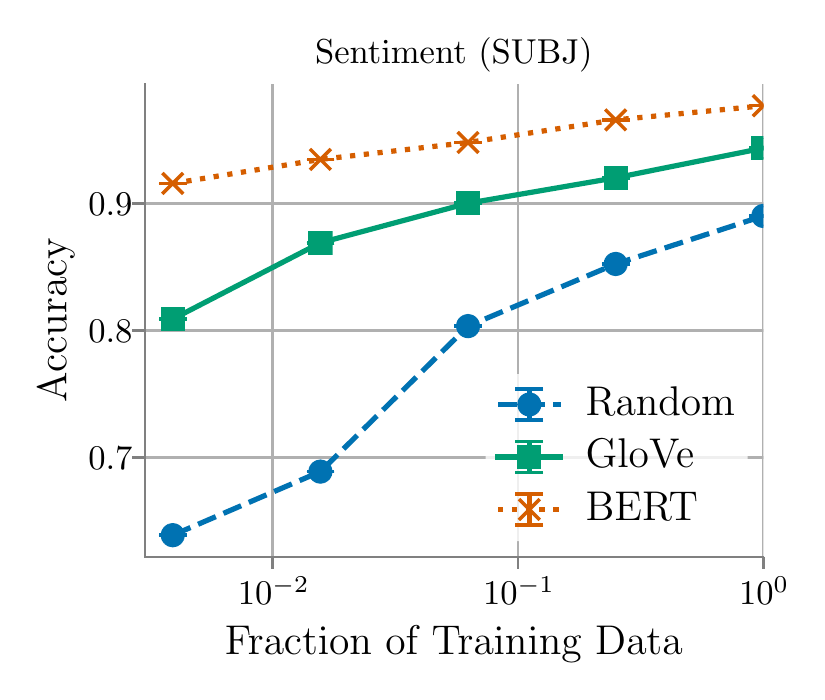} &
		\includegraphics[width=0.45\linewidth]{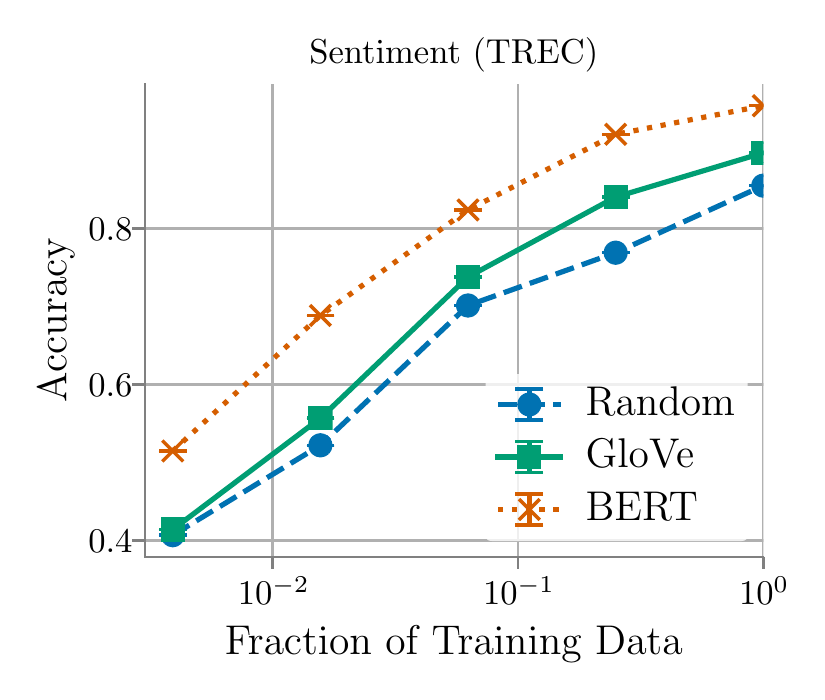}
	\end{tabular}
	\caption{Performance of random, GloVe, and BERT embeddings on the NER (top row) and sentiment analysis (bottom three rows) tasks as we vary the amount of training data \hspace{\textwidth}}
	\label{fig:sentiment_ner_vary_percent}
\end{figure*}

\begin{figure*}
	\begin{tabular}{c c}
		\includegraphics[width=0.45\linewidth]{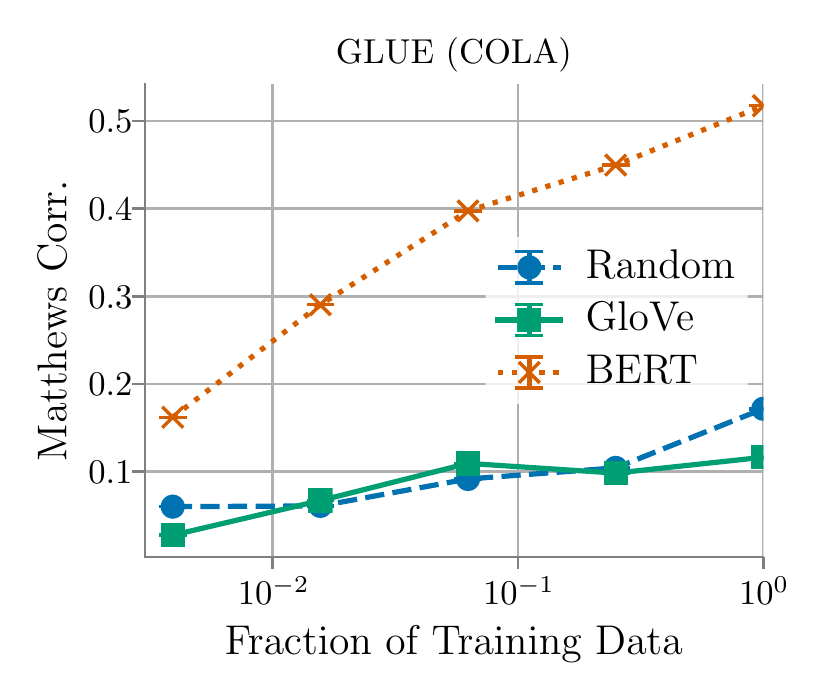} &
		\includegraphics[width=0.45\linewidth]{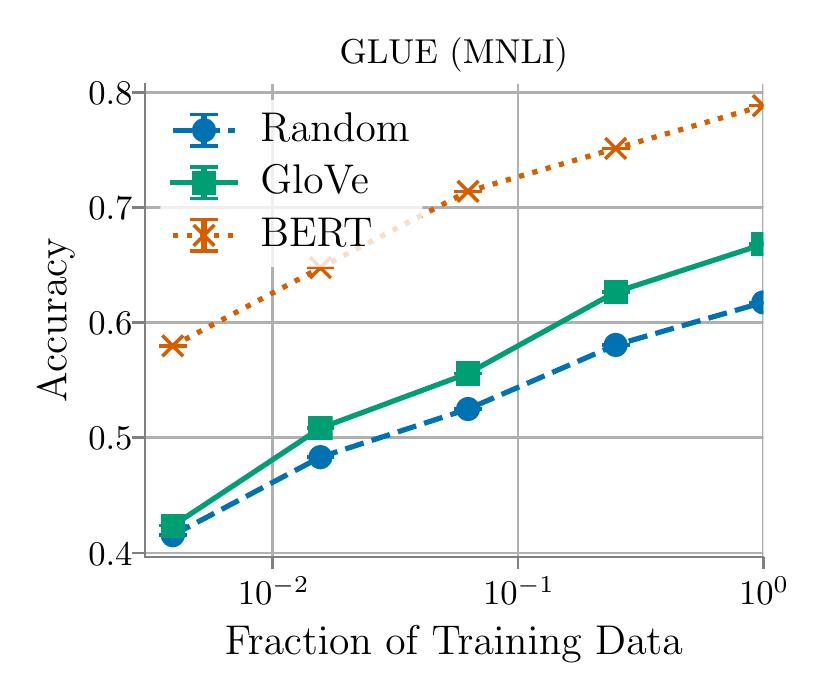} \\
		\includegraphics[width=0.45\linewidth]{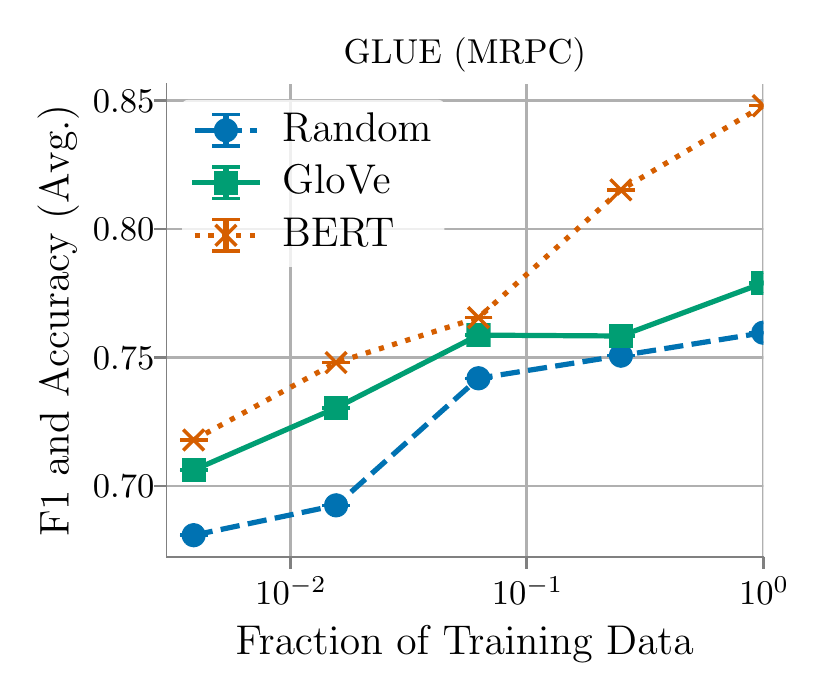} &
		\includegraphics[width=0.45\linewidth]{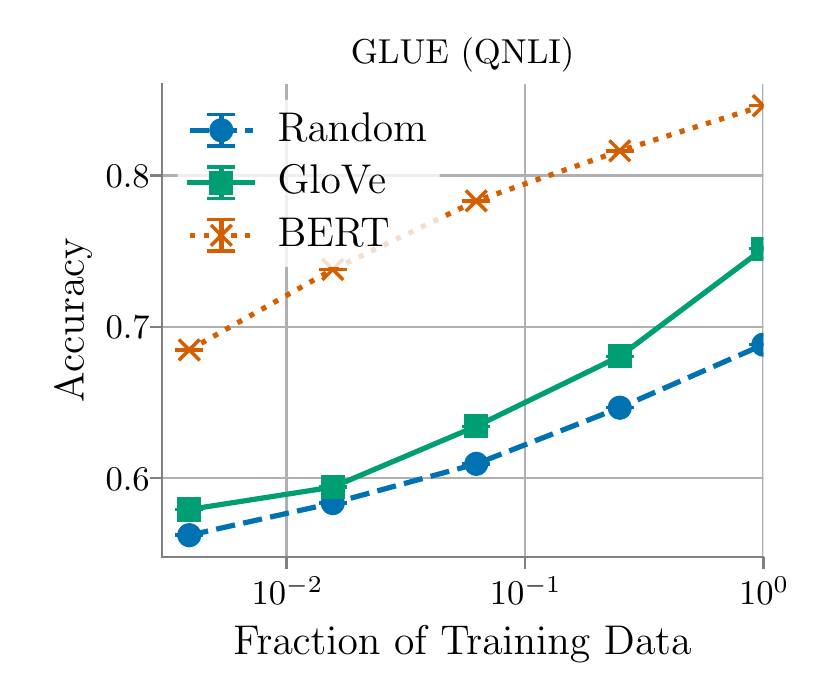} \\
		\includegraphics[width=0.45\linewidth]{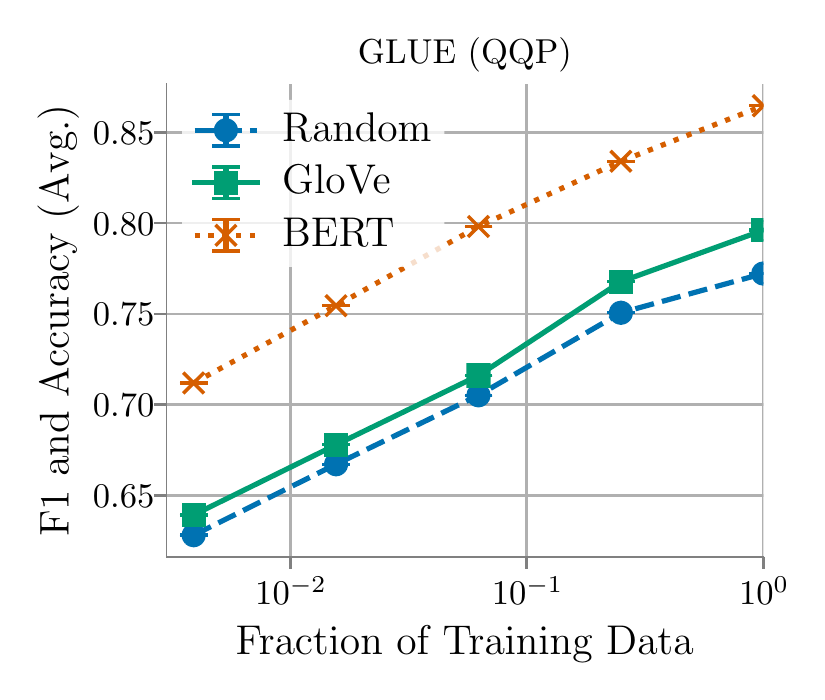} &
		\includegraphics[width=0.45\linewidth]{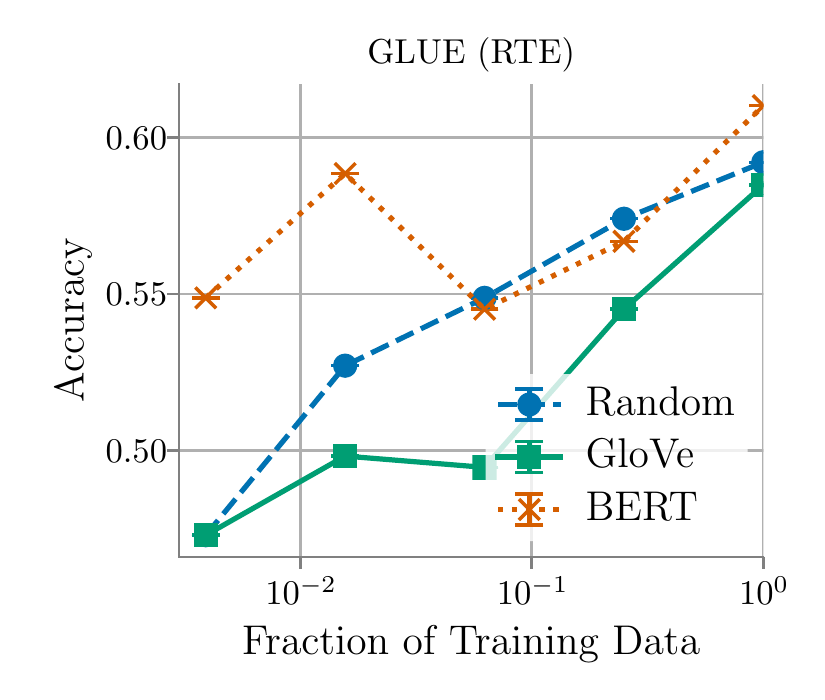} \\
		\includegraphics[width=0.45\linewidth]{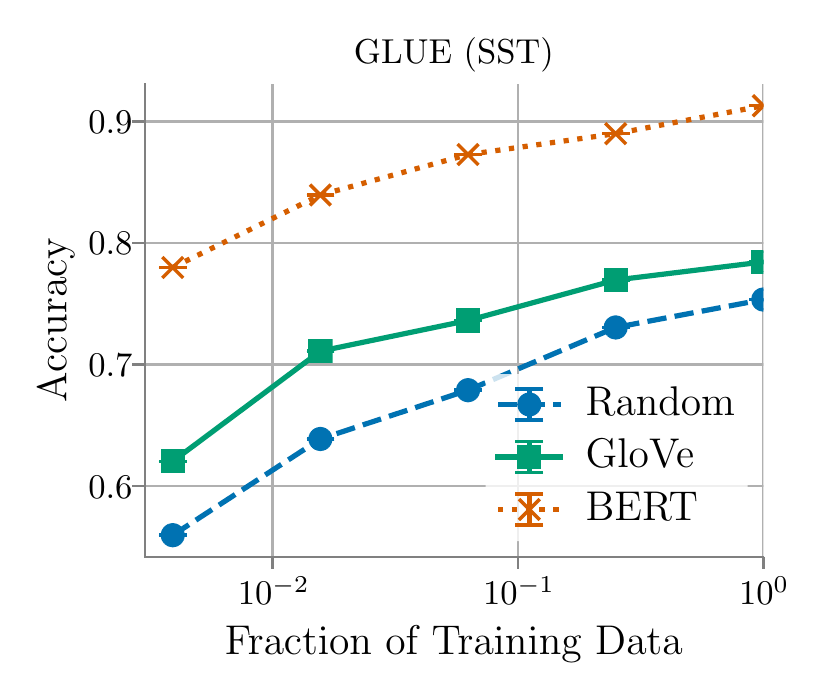} &
		\includegraphics[width=0.45\linewidth]{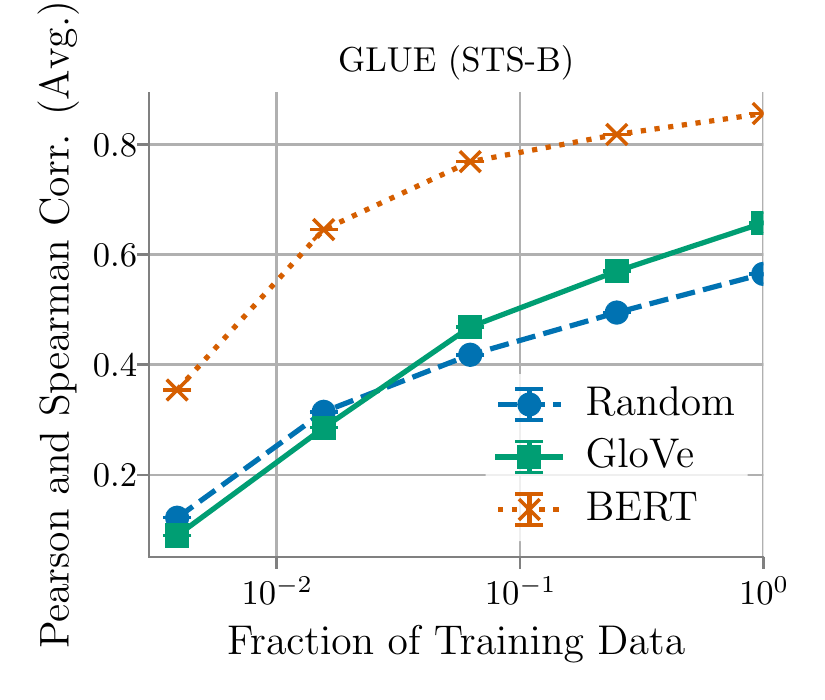}
	\end{tabular}
	\caption{Performance of random, GloVe, and BERT embeddings on GLUE tasks as we vary the amount of training data \hspace{\textwidth}}
	\label{fig:jiant_vary_percent}
\end{figure*}

In Figures~\ref{fig:sentiment_ner_vary_percent} and~\ref{fig:jiant_vary_percent}, we show the performance of random, GloVe, and BERT embeddings on all the NER, sentiment analysis, and GLUE tasks, as we vary the amount of training data.
We can see that across most of these results:
\begin{itemize}
	\item Non-contextual embedding performance improves quickly as the amount of training data is increased.
	\item The gap between contextual and non-contextual embeddings often shrinks as the amount of training data is increased.
	\item There are many tasks for which random and GloVe embeddings perform relatively similarly to one another.
\end{itemize}

\subsection{Theoretical Support for Random Embedding Performance}
\label{app:theory}
To provide theoretical support for why, given sufficient training data, a model trained with random embeddings might match the performance of one trained with pretrained embeddings, we consider the simple setting of Gaussian process (GP) regression \citep{gp06}.
In particular, we assume that the prior covariance function for the GP is determined by the pretrained embeddings, and show that as the number of observed samples from this GP grows, the posterior distribution gives diminishing weight to the prior covariance function, and eventually depends solely on the observed samples.
Thus, if we were to calculate the posterior distribution using an \textit{inaccurate} prior covariance function determined by random embeddings, this posterior would approach the true posterior as the number of observed samples grew.

More formally, for a fixed set of words $\{w_1,\ldots,w_n\}$ with pretrained embeddings $\{x_1,\ldots,x_n\} \subset \RR^d$, we assume that the ``true'' regression label vector $y^* \in \RR^n$ for these words is sampled from a zero-mean multivariate Gaussian distribution $y^* \sim \cN(0,K)$, where the entries $K_{ij} \defeq k(x_i,x_j)$ of the covariance matrix $K$ are determined based on the similarity $k(x_i,x_j)$ between the pretrained embeddings $x_i, x_j \in \RR^d$ for words $i$ and $j$.\footnote{As an example, we could have $k(x_i,x_j) \defeq \exp\left(-\|x_i-x_j\|^2/(2\sigma^2)\right)$ be the Gaussian kernel.}
We then assume that we observe $m$ noisy samples $(y_1,\ldots,y_m)$ of the ``true'' label vector $y^*$, where each $y_i\in \RR^n$ is an independent sample from $\cN(y^*, \sigma^2 I)$.  To summarize:
\begin{eqnarray*}
	y^* \!\!\!&\sim&\!\!\! \cN(0,K), \\
	y_1,\ldots,y_m \!\!\!&\sim&\!\!\! \cN(y^*, \sigma^2 I).
\end{eqnarray*}

The question then becomes, what is the posterior distribution for $y^*$ after observing $(y_1,\ldots,y_m)$?  The closed form solution for this posterior is as follows:
\begin{eqnarray*}
	p(y^* \!\!\!\!\!&|&\!\!\!\!\! y_1,\ldots,y_m) \;\;=\;\; \cN(\by_m,\bar{K}_m), \;\;\text{where}\\
	\by_m &=& K\left(K+\frac{\sigma^2}{m}I\right)^{-1}\left(\frac{1}{m}\sum_{i=1}^m y_i\right), \\
	\bar{K}_m &=& K\left(K+\frac{\sigma^2}{m}I\right)^{-1}\frac{\sigma^2}{m}I.
\end{eqnarray*}

Importantly, we observe that as $m\rightarrow \infty$, that $\by_m \rightarrow  y^*$ (because $K(K+\frac{\sigma^2}{m}I)^{-1} \rightarrow I$ and $\frac{1}{m}\sum_{i=1}^m y_i \rightarrow y^*$), and $\bar{K}_m \rightarrow 0$.
Thus, if we were to compute the posterior distribution for this GP using an uninformative prior covariance function $K'$ determined by random embeddings $\{x_1',\ldots,x_n'\}$  ($K'_{ij} = k(x_i',x_j')$), this posterior would approach the posterior computed from the ``true'' prior covariance function $K$ as the number of observations $m\rightarrow \infty$.
\textit{Thus, GP regression with an informative prior derived from the pretrained embeddings performs the same as GP regression with an uninformative prior derived from random embeddings, as the number of observed samples approaches infinity}.

\section{Study of Linguistic Properties}
\label{app:linguistic}
We now describe in more detail how we define our metrics for the three linguistic properties for both NER and sentiment analysis tasks (Appendix~\ref{app:linguistic_definitions}), as well as provide extended empirical results from our linguistic studies (Appendix~\ref{app:extended_results_linguistic}).

\subsection{Linguistic Properties: Detailed Definitions}
\label{app:linguistic_definitions}

\begin{figure*}[htb!]
	\includegraphics[width=\linewidth]{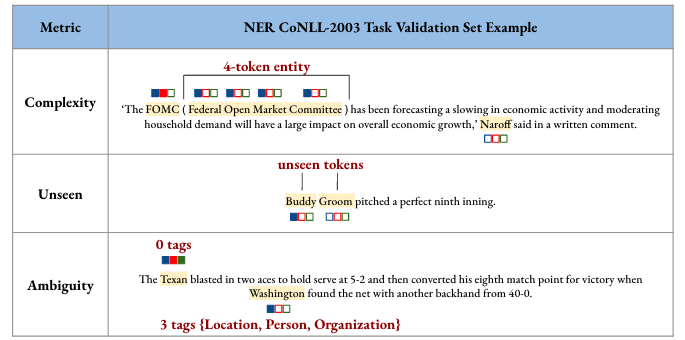}
	\includegraphics[width=\linewidth]{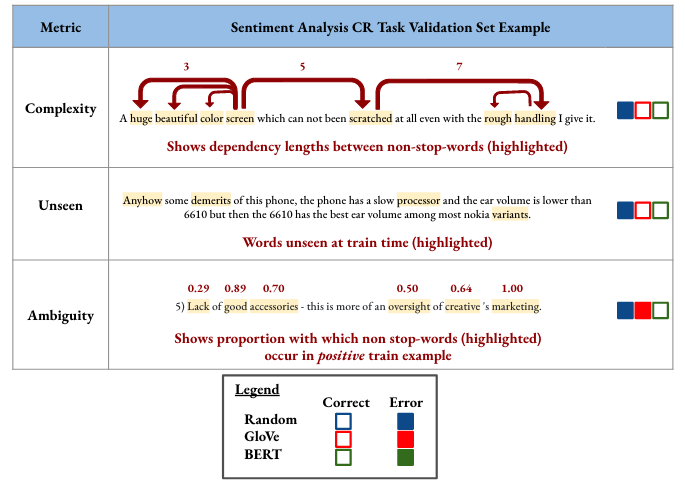}
	\caption{Examples from the CoNLL-2003 NER task (above) and the CR sentiment analysis task (below) validation sets, to provide further intuition for the three linguistic properties.
	All of the examples above fall in the validation set slices that have metric values above the median, and are thus considered relatively difficult examples according to these linguistic metrics.
	For example, in the case of NER, (1) the ``Federal Open Market Committee'' is a relatively long, 4-token entity, (2) ``Buddy'' and ``Groom'' are both tokens that were not seen during training, and (3) ``Washington'' was seen in the training set with three different entity type labels (location, person, organization).
	In the case of the sentiment analysis examples, (1) the complexity metric sentence has several long dependences (lengths 3, 5, and 7) because it has numerous adjective, adverb, and noun modifiers, (2) the unseen metric sentence has four words that were not seen during training (``anyhow'', ``demerits'', ``processor'', ``variants''), and (3) the ambiguity metric sentence has words that were mainly positive during training (``good'', ``creative''), as well as words which were mainly negative during training (``lack'').
	We use empty vs.\ filled-in squares of different colors to show whether a given embedding type got an example correct vs.\ incorrect, respectively (see legend).
	\hspace{\textwidth}}
	\label{fig:ner_examples}
\end{figure*}

We define the metrics in detail below for our three linguistic properties: complexity of text structure (Appendix~\ref{app:complexity}), ambiguity in word usage (Appendix~\ref{app:ambiguity}), and prevalence of unseen words (Appendix~\ref{app:unseen}).
To provide further intuition for these metrics, in Figure~\ref{fig:ner_examples} we present actual examples from the CoNLL-2003 NER task and the CR sentiment analysis task for each of the metrics, along with the errors made by each embedding type on these examples.

\subsubsection{Complexity of Text Structure}
\label{app:complexity}
We define the following metrics for NER and sentiment analysis to measure the structural complexity of an entity or sentence, respectively:

\textbf{NER}: For NER, we measure the linguistic complexity of an entity in terms of the number of tokens in the entity (\eg, ``George Washington'' spans 2 tokens), as correctly labeling a longer entity requires understanding the \textit{relationships} between the different tokens in the entity name.

\textbf{Sentiment analysis}: For sentiment analysis, we need a sentence-level proxy for structural complexity; toward this end, we leverage the dependency parse tree for each sentence in the dataset.\footnote{We use the StanfordNLP dependency parser for our metric: \url{https://pypi.org/project/stanfordnlp/}.} In particular, we characterize a sentence as more structurally complex if the average distance between dependent words is higher.  
We consider this definition because long-range dependencies generally require more contextual information to understand.
To avoid diluting the average dependency length, we do not include dependencies where either the head or the tail of the dependency is a punctuation or a stop word.

As an example, consider the sentence ``George Washington, who was the first president of the United States, was born in 1732''.
In this sentence, there is a dependence between ``George'' and ``born'' of length 14, because there are 13 intervening words or punctuations.
This is a relatively large gap between dependent words, and would increase the average dependency length the sentence.

\subsubsection{Ambiguity in Word Usage}
\label{app:ambiguity}
The next linguistic property we consider is the degree of ambiguity in word usage within a task.
To measure the degree of ambiguity in the language, we define the following metrics in the context of NER and sentiment analysis:

\textbf{NER}: For NER as a word-level classification task, we consider the number of labels (person, location, organization, miscellaneous, other) a token appeared with in the training set as a measure of its ambiguity (\eg, ``Washington'' appears as a person, location, and organization in the CoNLL-2003 training set).
For each token in the validation set, we enumerate the number of tags it appears with in the training set. 

\textbf{Sentiment analysis}: For sentiment analysis, we measure the ambiguity of a sentence by considering whether the words in the sentence generally appear in positive or negative sentences in the training data.
For the binary case, we take the average over words in the sentence of the unigram probability that a word is positive, and then compute the entropy of a coin flip with this probability of being ``heads''.
More specifically, to compute the unigram probability $p(+1\,|\,w)$ for a word $w$, we measure the fraction of training sentences containing $w$ which are positive.
Our ambiguity metric is then defined for a sentence $S$ as
$$H\left(\frac{1}{|S|}\sum_{w \in S} p(+1\,|\,w)\right),$$ 
where $H(p) = -p\log_2(p) - (1-p)\log_2(1-p)$ is the entropy of a coin flip with probability $p$.
Intuitively, sentences with generally positive (or negative) words will have low entropy, and be easy to classify even with non-contextual embeddings.

For non-binary sentiment tasks with $C$-labels (\eg, $C=6$ for the TREC dataset), we consider the entropy of the average label distribution $\frac{1}{|S|}\sum_{w\in S} p(y\,|\,w) \in \RR^C$ over the words in the sentence.
Here, $p(y\,|\,w)$ is defined as the fraction of the sentences in the training set containing the word $w$ which had the label $y$.
Note that for stop words and punctuation, we always consider $p(y\,|\,w)$ as the uniform distribution over the set of possible labels $y$ (for both binary and non-binary classification tasks).

\begin{table}	
	\centering	
	\begin{tabular}{cccc}	
		\hline \textbf{Category} & \textbf{BERT} & \textbf{Random} & \textbf{GloVe} \\ \hline	
		LS  & 0.19         & 0.14         & 0.13 \\	
		PAS & 0.33         & 0.20         & 0.20 \\	
		L   & 0.12         & 0.15         & 0.13 \\	
		KCS   & 0.10         & 0.17         & 0.13 \\	
		Overall   & 0.500		  & 0.475  & 0.465 \\	
		\hline	
	\end{tabular}	
	\caption{\label{tab:glue_diagnostics} The performance (Matthews correlation coefficients) of BERT, random, and GloVe embeddings across the four linguistic categories defined by the GLUE diagnostic task: lexical semantics (LS), predicate-argument structure (PAS), logic (L), and knowledge and common sense (KCS). We also include the overall diagnostic performance.}	
\end{table}

\subsubsection{Prevalence of Unseen Words}
\label{app:unseen}
We define the following metrics for the prevalence of unseen words for NER and sentiment analysis tasks:

\textbf{NER}: For a word in the NER validation set, we consider as our metric the inverse of the number of times the word appeared in the training data (letting $1/0\defeq \infty$).
We consider the \textit{inverse} of the number of training set appearances because intuitively, if a word appears \textit{fewer} times in the training set, we expect it to be \textit{harder} to correctly classify this word at test time---especially for non-contextual or random embeddings.

\textbf{Sentiment analysis}: For sentiment analysis, given a sentence, we consider as our metric the fraction of words in the sentence that were \textit{never} seen during training.
More specifically, we count the number of unseen words (that are not stop words), and divide by the total number of words in the sentence.
Intuitively, sentences with many unseen words will attain high values for this metric, and will be difficult to classify correctly without prior (\ie, pretrained) knowledge about these unseen words.

\subsection{Extended Results}
\label{app:extended_results_linguistic}

\begin{table*}
	\centering	
	\begin{tabular}{l l l l l l l}	
		\hline \multirow{2}{*}{} & \multicolumn{2}{c}{\textbf{Complexity}} & \multicolumn{2}{c}{\textbf{Ambiguity}} & \multicolumn{2}{c}{\textbf{Unseen}} \\	
		\cmidrule(lr){2-3}	
		\cmidrule(lr){4-5}	
		\cmidrule(lr){6-7}	
		\textbf{Task} & Abs.\!\!\!\! & Rel.\!\!\!\! & Abs.\!\!\!\! & Rel.\!\!\!\! & Abs.\!\!\!\! & Rel.\!\!\!\! \\ \hline	
		NER (CoNLL) & +6.7 & 4.0 & +5.9 & 3.3 & -1.4 & 0.8 \\	
		Sent. (MR) & -0.6 & 0.9 & +6.5 & 2.5 & -1.0 & 0.9 \\	
		Sent. (SUBJ) & -1.8 & 0.6 & +4.4 & 6.0 & -1.3 & 0.6 \\	
		Sent. (CR) & +1.2 & 1.5 & -2.4 & 0.4 & 0.0 & 1.0 \\	
		Sent. (SST) & +7.8 & 5.3 & +6.0 & 3.2 & -2.8 & 0.6 \\	
		Sent. (TREC) & +2.2 & 1.4 & +8.1 & 4.1 & +3.7 & 1.8 \\	
		Sent. (MPQA) & +6.6 & -3.2 & +2.9 & -1.8 & +0.4 & 3.0 \\	
		\hline	
	\end{tabular}	
	\caption{\label{tab:metrics_glove} For our complexity, ambiguity, and unseen prevalence metrics, we slice the validation set using the median metric value, and compute the average error rates for GloVe and BERT on each slice.	
		We show that the gap between GloVe and BERT errors is larger above than below the median in 11 out of 14 of the complexity and ambiguity results both in absolute (Abs.) and relative (Rel.) terms; however, on the unseen metrics, this only holds for 2 out of 7 cases, which suggests that GloVe embeddings are able to relatively effectively deal with unseen words. 	
	}	
\end{table*}

We present the detailed results from our evaluation of the different embedding types on the GLUE diagnostic dataset (Appendix~\ref{app:glue_diagnostic}), and extended validation of the linguistic properties we define in Section~\ref{sec:linguistics} (Appendix~\ref{app:linguistic_glove_vs_bert}).

\subsubsection{GLUE Diagnostic Results}
\label{app:glue_diagnostic}
The GLUE diagnostic task facilitates a fine-grained analysis of a model's strengths and weaknesses in terms of how well the model handles different linguistic phenomena.
The task consists of 550 sentence pairs which are classified as entailment, contradiction, or neutral.
The GLUE team curated the sentence pairs to represent over 20 linguistic phenomena, which are grouped in four top-level categories: lexical semantics (LS), predicate-argument structure (PAS), logic (L), and knowledge and common sense (KCS).
We follow the standard procedure and use the model trained on the MNLI dataset (using the random, GloVe, or BERT embeddings) to evaluate performance on the diagnostic task.
We report the Matthews correlation coefficient (MCC) performance of the different embedding types on the four top-level categories in Table~\ref{tab:glue_diagnostics}.

Our two key observations are:
(1) the non-contextual embeddings (random and GloVe) perform similarly to one another across all four top-level categories;
(2) the performance difference between contextual and non-contextual embeddings is most stark for the predicate-argument (PAS) category, which includes phenomena that require understanding the interactions between the different subphrases in a sentence.
Within PAS, the BERT embeddings attain a 10+ point improvement in MCC over random embeddings for sentences reflecting the following phenomena: Relative Clauses/Restrictivity, Datives, Nominalization, Core Arguments, Core Arguments/Anaphora/Coreference, and Prepositional Phrases.

\subsubsection{GloVe vs.\ BERT Results}
\label{app:linguistic_glove_vs_bert}
In Table~\ref{tab:metrics_glove}, we replicate the results from Table~\ref{tab:metrics}, but instead of comparing BERT embeddings to random embeddings, we compare them to GloVe embeddings.
We can see that for 11 out of 14 cases for the complexity and ambiguity metrics, the gap between contextual (BERT) and non-contextual (GloVe) performance is larger for the validation slices above the median than below; this aligns with our results comparing random and BERT embeddings.
Interestingly, this is only the case for 2 out of 7 of the cases for the unseen metrics.
This is likely because both GloVe and BERT embeddings are able to leverage pretrained semantic information about unseen words to make accurate predictions for them, and thus perform relatively similarly to one another on unseen words.